\pdfoutput=1
\documentclass[preprint,3p,12pt]{elsarticle}

\usepackage{amssymb}
\usepackage{graphicx}
\usepackage{subfigure}
\usepackage{multirow}
\usepackage{booktabs}
\usepackage{hyperref}
\journal{Expert Systems with Applications}

\begin{document}

\begin{frontmatter}



\title{X2-Softmax: Margin Adaptive Loss Function for Face Recognition}

\author[address1]{Jiamu Xu}
\ead{xujiamu@stu2022.jnu.edu.cn}

\author[address1]{Xiaoxiang Liu}
\ead{tlxx@jnu.edu.cn}
\author[address1]{Xinyuan Zhang}
\ead{Zhangxy@jnu.edu.cn}
\author[address3]{Yain-Whar Si}
\ead{fstasp@um.edu.mo}
\author[address1]{Xiaofan Li}
\ead{lixiaofan@jnu.edu.cn}
\author[address1]{Zheng Shi}
\ead{zhengshi@jnu.edu.cn}
\author[address2]{Ke Wang}
\ead{wangke@jnu.edu.cn}
\author[address1]{Xueyuan Gong\corref{cor1}}
\ead{xygong@jnu.edu.cn}
\cortext[cor1]{Corresponding authors.}
\affiliation[address1]{organization={School of Intelligent Systems Science and Engnieering},
             addressline={Jinan University},
             state={Guangdong Province},
             country={China}}
\affiliation[address2]{organization={College of Information Science and Technology},
             addressline={Jinan University},
             state={Guangdong Province},
             country={China}}
\affiliation[address3]{organization={Department of Computer and Information Science},
             addressline={Faculty of Science and Technology, University of Macau},
             state={Macau},
             country={China}}

%\author[label1,label2]{}
%\affiliation[label1]{organization={},
  %           country={}}
%%
%\affiliation[label2]{organization={},

% \author[label1]{Jiamu Xu, Xiaoxiang Liu, Xinyuan Zhang, Yain-Whar Si, Xiaofan Li, Zheng Shi, Ke Wang, Xueyuan Gong}
% \affiliation[label1]{organization={Jinan University},
%            country={China}}

\begin{abstract}
Learning the discriminative features of different faces is an important task in face recognition. 
By extracting face features in neural networks, it becomes easy to measure the similarity of different face images, which makes face recognition possible.
To enhance the neural network's face feature separability, incorporating an angular margin during training is common practice.
State-of-the-art loss functions CosFace and ArcFace apply fixed margins between weights of classes to enhance the inter-class separation of face features. Since the distribution of samples in the training set is imbalanced, similarities between different identities are unequal. 
Therefore, using an inappropriately fixed angular margin may lead to the problem that the model is difficult to converge or the face features are not discriminative enough.
It is more in line with our intuition that the margins are angular adaptive, which could increase with the angles between classes growing.
In this paper, we propose a new angular margin loss named X2-Softmax. 
X2-Softmax loss has adaptive angular margins, which provide the margin that increases with the angle between different classes growing.
The angular adaptive margin ensures model flexibility and effectively improves the effect of face recognition. 
We have trained the neural network with X2-Softmax loss on the MS1Mv3 dataset and tested it on several evaluation benchmarks to demonstrate the effectiveness and superiority of our loss function. 
\end{abstract}




\end{frontmatter}


\section{Introduction}
\label{sec:intro}
Face recognition is being widely used in situations where identity verification is required, such as access control management and application verification services~\cite{arindam2022rbeca}. 

Face recognition has four steps: face detection, face alignment, face feature extraction, and feature comparison~\cite{wang2022survey, chaorong2019dependence}. 
Face feature extraction is one of the most important tasks in the face recognition process. 
To improve the accuracy of face recognition, it is necessary to enhance the model's ability to extract discriminative face features.

The loss function for extracting face features can usually be divided into two categories. 
One category is pair-based loss (e.g., Contrastive loss~\cite{chopra2005learning}, Triple loss~\cite{schroff2015facenet, hoffer2015deep}, N-pairs loss~\cite{sohn2016improved}), and the other one is classification-based loss (e.g., Softmax loss~\cite{sun2014deep}, CosFace~\cite{wang2018cosface}, ArcFace~\cite{deng2019arcface}).
However, training with pair-based loss, the computational time will increase significantly with the growing number of sample pairs in training dataset~\cite{liu2016large,boutros2022elasticface}, and redundant pairs may lead to a slowly converging and degenerated model~\cite{Wang2019multisimilarity}.

\begin{figure}[!ht]
    \centering
    \subfigure[fixed angular margins]{
        \begin{minipage}[t]{0.4\linewidth}
            \centering
			\includegraphics[width=1\linewidth]{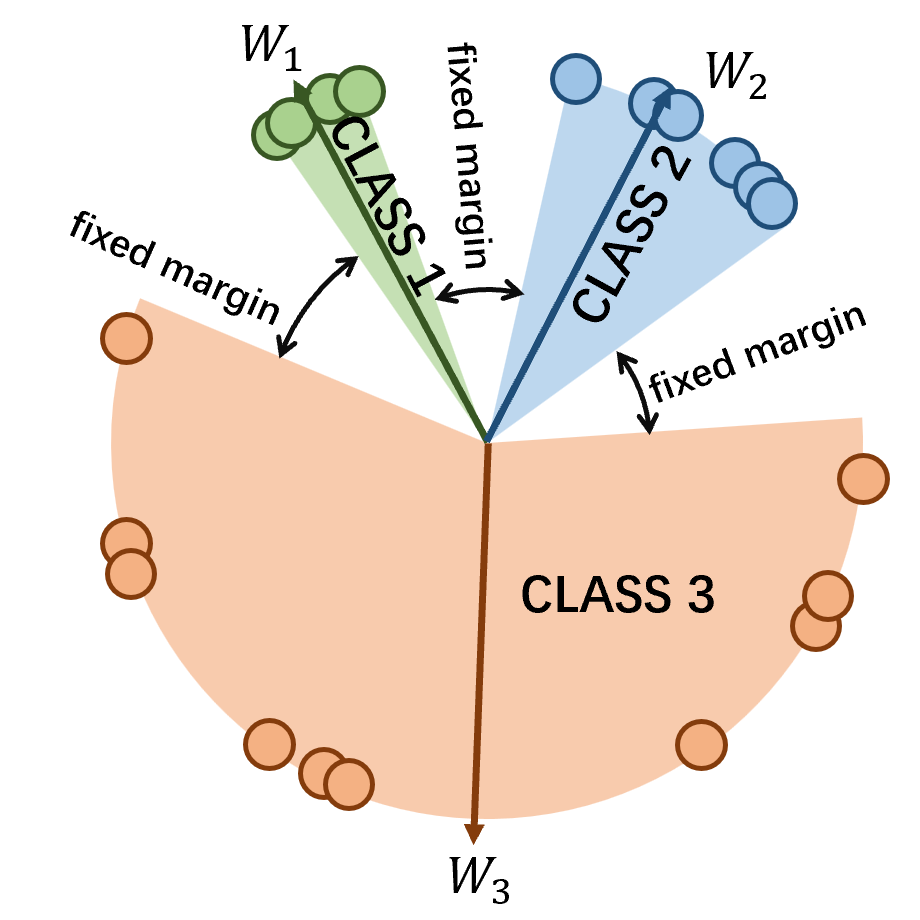}
            \label{sfig:fix_m}
        \end{minipage}
    }
    \subfigure[adaptive angular margins]{
        \begin{minipage}[t]{0.4\linewidth}
            \centering
			\includegraphics[width=1\linewidth]{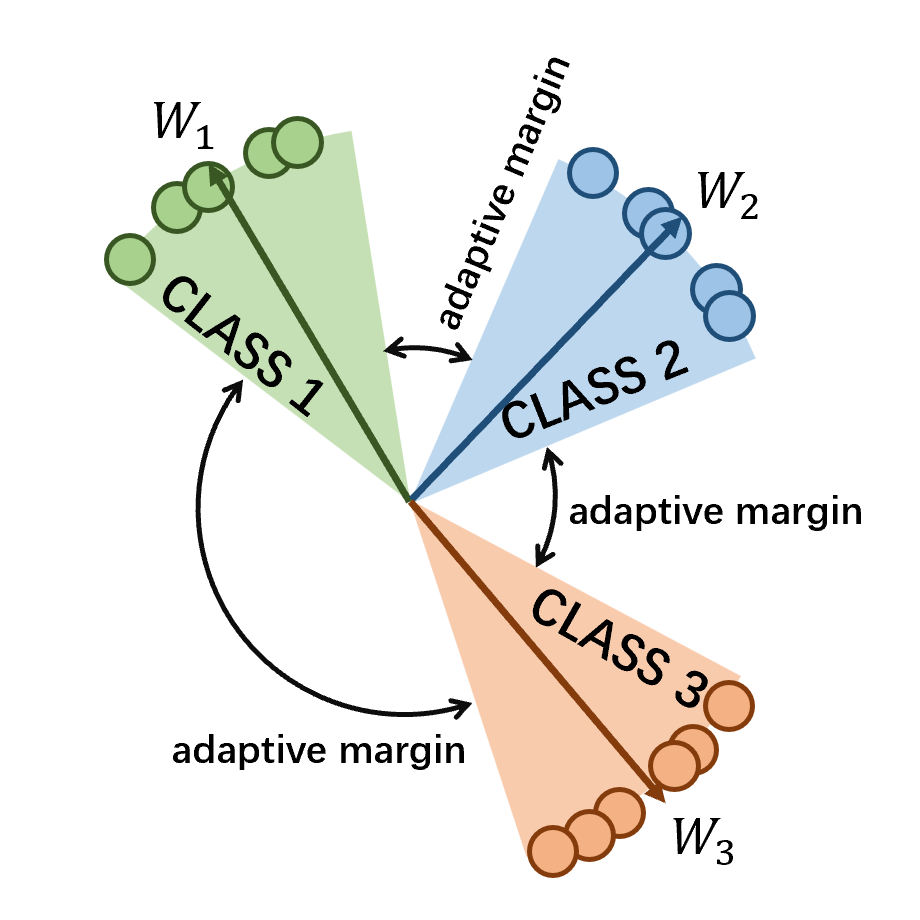}
            \label{sfig:ada_m}
        \end{minipage}
    }
    \caption{Since the distribution of face training samples is imbalanced, using a fixed-size margin would not better fit all imbalanced distributed samples than using an adaptive margin.}
\end{figure}

The Softmax loss function is commonly used in classification tasks, and face recognition can also be regarded as a classification task.
However, the face features extracted with Softmax loss are not discriminative enough for the open-set face recognition problem~\cite{deng2019arcface}.
To enhance intra-class compactness and inter-class separation, people incorporated different fixed angular margins in Softmax loss, such as CosFace~\cite{wang2018cosface} and ArcFace~\cite{deng2019arcface}. 
While face features become more discriminative, choosing an optimal angular margin becomes a new problem. 
As shown in Fig.~\ref{sfig:fix_m}, since the samples in the training set are unevenly distributed, it is not appropriate to add a fixed angular margin between classes.
The model will be difficult to converge if the fixed angular margin is too large~\cite{boutros2022elasticface}. 
On the contrary, when the fixed angular margin is too small, its contribution to feature separation is minimal.
What's more, training with a fixed angular margin may cause neural networks to overfitting~\cite{jiao2021dyn}.

From a practical point of view, we want the margin to be adaptive, and able to change its size with the angle between different classes as shown in Fig.~\ref{sfig:ada_m}. 
For two classes with larger angles, we want them to have a larger margin to promote the enhancement of intra-class compactness. 
For two classes with smaller angles, we would like a proper margin that increases the inter-class separation without making it too large for the model to converge.

In recent years there have also been several loss functions~\cite{jiao2021dyn, meng2021magface} with adaptive angular margins to enhance intra-class compactness and inter-class separation of extracted face features.
However, Dyn-ArcFace~\cite{jiao2021dyn} adds dynamic angular margins calculated by inter-class center distances to the nearest neighbor class.
The angular margins will become small in Dyn-ArcFace when the distance between two classes is close, which will have little effect in these two classes to other classes with larger distances.
The angular margins in MagFace~\cite{meng2021magface} are related to modules of features rather than the similarity between classes. 
It still doesn't solve the problem that classes with larger angles should be set at a larger margin.
% \begin{figure}[h]
%    \centering
%    \includegraphics[width=0.5\linewidth]{1_dyn_margin.png}
%    \caption{In Dyn-ArcFace loss, the margin between class to other classes depends on the distance to the nearest neighbor class, which in the figure is class 2.
%    The margins between class 1 to other classes, such as with class 5, will also be determined by the distance to the nearest neighbor class, which limits the enhancement of inter-class separation of face features.}
%    \label{fig:dyn_margin}
% \end{figure}

In this paper, we present X2-Softmax loss, which replaces the cosine function in ArcFace loss by using a quadratic function. 
Since the X2-Softmax loss is margin-adaptive, it can automatically adjust the angular margin with the angle between weights of classes. 
To demonstrate the effectiveness and superiority of our X2-Softmax loss, we tested it on eight different test benchmarks and compared the results with other loss functions. 
From the experimental results, X2-Softmax loss achieves promising and competitive results on most benchmarks.
% TODO: Experiment Results

In summary, the main contributions of this paper can be concluded as follows:
\begin{itemize}
	\item A loss function named X2-Softmax is proposed. 
    To solve the problem that difficult to choose an ideal fixed angular margin for existing loss functions, an adaptive angular margin loss function is proposed for face recognition. 
    Different from the traditional loss function, the X2-Softmax loss function has an adaptive margin, which could avoid choosing a suitable fixed margin to accommodate differences in the distribution of different classes.
	\item Our X2-Softmax loss function was tested on eight different test benchmarks and achieved promising results, which proves its effectiveness and superiority. 
    The experimental results prove that X2-Softmax loss function obtains competitive performance.
\end{itemize}
% TODO: Experiment Results

The rest of the paper is structured as follows.
In Section 2, the existing related work on loss functions is presented.
In Section 3, our proposed X2-Softmax loss function is presented.
In Section 4, our experimental setup and experimental results are presented.
In Section 5, we discuss the results obtained and conclude.

\section{Related work}
\label{sec:rw}
Currently, there are two main directions of loss function for face feature extraction.
One is pair-based loss, and the other is classification-based loss.

Pair-based loss learns face features directly by using pairs of samples.
Contrastive loss~\cite{hadsell2006dimensionality} is one of the most straightforward pair-based losses.
It maps face features into Euclidean space by ensuring the Euclidean distance of positive samples is smaller than negative samples with a fixed margin.
Similarly, Cosine Contrastive loss~\cite{Kelong2021SimpleX} maps face features by enforcing the cosine distance of sample pairs rather than Euclidean distance.
Triplet loss~\cite{schroff2015facenet} trains with triple pairs of anchor samples and their positive samples and negative samples. 
The Euclidean distance between the face features of the anchor and its negative sample is forced to be larger than the anchor to its positive sample.
Triplet center loss~\cite{he2018triplet} considers the Euclidean distance from anchor to positive class center and negative class center instead of positive sample and negative sample. 
Sohn proposes the N-pair loss~\cite{sohn2016improved} to solve the problem of slow convergence by learning $N-1$ negative samples at the same time.
Pair-based loss directly learns to face features embedding into the feature space with pairs of samples and enhances intra-class compactness and inter-class separation. 
However, the computational time will increase significantly while the number of images in the training set grows~\cite{liu2016large,boutros2022elasticface}, and mining half-hard pairs becomes more difficult.
What's more, the choice of sample pairs used for training affects the results of model training. 
Random sampling can lead to a huge number of redundant training pairs, which results in slower convergence and model degradation during training, thus further reducing the efficiency of training~\cite{Wang2019multisimilarity}.

Unlike pair-based loss, classification-based loss aims to extract face features by implementing the classification task.
Classification-based loss does not encounter the problem of training data increasing explosively with the number of images in the training set.
Softmax loss is a typical classification-based loss, which is widely used in classification tasks~\cite{krizhevsky2012imagenet}.
The face recognition task can also be seen as a classification problem, and thus softmax loss is also commonly used in face recognition.
As the simplest classification-basked loss, softmax loss can compact face features of the same class, and separate face features of different classes while completing the classification task. 
Since softmax loss does not directly optimize the extraction of face features, it is not accurate enough in facing the open-set face recognition problem~\cite{deng2019arcface}. 
To improve the accuracy of face recognition, some loss functions are derived based on softmax loss.

L-Softmax~\cite{liu2016large} and SphereFace(A-Softmax)~\cite{liu2017sphereface} build on the original softmax loss function by adding a multiplicative angular margin, which promotes the generation of features with a larger inter-class variation. 
The difference is that SphereFace normalizes the weights of the last fully connected layer at the same time, making the classification based on angles instead of vector inner products.
The multiplicative angular margin loss represented by L-Softmax and SphereFace has a steep target logit curve which is difficult to converge and increases the difficulty of model training.
By normalizing the weights and face features, Wang et al propose NormFace~\cite{wang2017normface} to alleviate data imbalance and directly optimize the cosine similarity.
AM-Softmax~\cite{wang2018additive} and CosFace~\cite{wang2018cosface} propose to add a fixed angular margin in cosine space to enable face features to become more discriminative.
Different from CosFace, ArcFace~\cite{deng2019arcface} adds a fixed angular margin in angle space instead of cosine space.
However, there are also some problems with training by loss with a fixed angular margin.
The fixed angular margin determines the accuracy of the neural network, but it is not an easy task to choose an ideal fixed margin.
In addition, the same fixed margin, which is suitable between some classes, may be too large to converge between other classes, or too small to promote significant intra-class compactness of face features between some other classes.
Since the similarity of face features across different classes of training samples is not consistent, it is not reasonable to add the same angular margin between classes with different similarities to enhance intra-class compactness and inter-class separation.

Dyn-ArcFace~\cite{jiao2021dyn} adds a dynamic angular margin, which is determined by the inter-class center distance to the nearest neighbor class.
The authors demonstrated that the dynamic angular margin will reduce the degree of overfitting during training by experiment.
The dynamic angular margins of Dyn-ArcFace depend on the inter-class center distances to the nearest neighbor class, which limits the margins to enhance the inter-class separation of face features to other classes.
MagFace~\cite{meng2021magface} adds a dynamic angular margin related to the module of the face features, which prevents model overfitting with noisy and low-quality samples.
The angular margins in MagFace are determined by the features module, but it still doesn't solve the problem of setting a larger angular margin for a larger angle between the weights of classes.
In ElasticFace~\cite{boutros2022elasticface}, the authors believe that the use of a fixed margin limits the flexibility of the training face recognition model.
Thus ElasticFace trains with dynamic angular margins, which are drawn from a Gaussian distribution instead of the fixed margins in CosFace and ArcFace loss.
ElasticFace+, an extension of ElasticFace, is also mentioned in the paper, which distributes the generated random margins according to the distance between the sample and other class centers. 
Generating random values in ElasticFace means that the generated margins are potentially redundant, which results in unnecessary computations.
Redundant margin generation is avoided in Ealstic+ by sorting. 
However, as the authors show in their paper, the increase in training computation due to sorting the margins is too huge to tolerate.
The adaptive margin in HAMFace~\cite{li2024hamface} is related by the angle between the face feature and ground truth weights.
But this also means that HAMFace ignores the effect of the similarity between different classes.

The X2-Softmax loss function proposed in this paper does not use a fixed margin but uses the function itself to obtain different margins with angles between the weights of two classes.
This method can circumvent the problem that it is difficult to choose a fixed margin for training because of the uneven distribution of the samples themselves.

\section{Proposed approaches}
\label{sec:pa}
In this section, we will introduce our loss function X2-Softmax, which aims to improve the accuracy of face recognition through its margin adaptivity.

\subsection{Preliminary}
\label{subsec:pre}
CosFace and ArcFace apply a fixed angular margin in cosine space and angle space based on Softmax loss function, respectively.
These two loss functions can be expressed by a similar formula as follows:

\begin{equation}
    \label{eq:basicfomula}
    L=-\frac{1}{N}\sum_{i=1}^{N}\log{\frac{e^{sf(\theta_{y_{i}})}}{e^{sf(\theta_{y_{i}})} +\sum_{j=1, j\ne y_{i}}^{C}e^{s\cos {\theta_{j}}}}}
\end{equation}

The difference between CosFace and ArcFace in Eq.~\ref{eq:basicfomula} is the logits function $f(\theta)$.
In CosFace, the logits function $f(\theta)$ is:
\begin{equation}
    \label{eq:cosf}
    f_{C}(\theta)=\cos{\theta}-m
\end{equation}

In ArcFace, the logits function $f(\theta)$ is:
\begin{equation}
    \label{eq:arcf}
    f_{A}(\theta)=\cos(\theta+m)
\end{equation}

For Eq.~\ref{eq:cosf} and Eq.~\Ref{eq:arcf} above, $m$ is the hyperparameter of the loss function, which represents the margins in cosine space and angle space.

\begin{figure}[ht]
    \centering
    \subfigure[Softmax]{
        \begin{minipage}[t]{0.3\linewidth}
             \centering
             \includegraphics[width=1\linewidth]{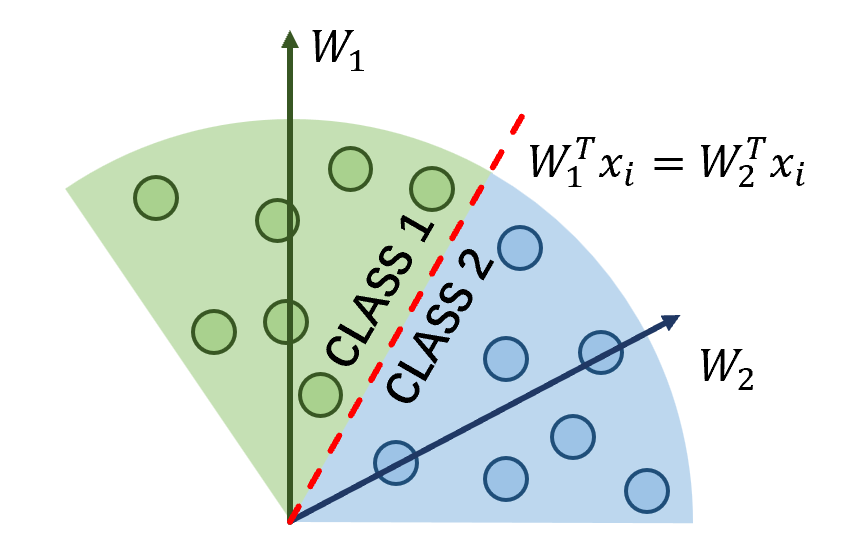}
            \label{sfig:dbSoftmax}
        \end{minipage}
    }
    \subfigure[CosFace]{
        \begin{minipage}[t]{0.3\linewidth}
             \centering
             \includegraphics[width=1\linewidth]{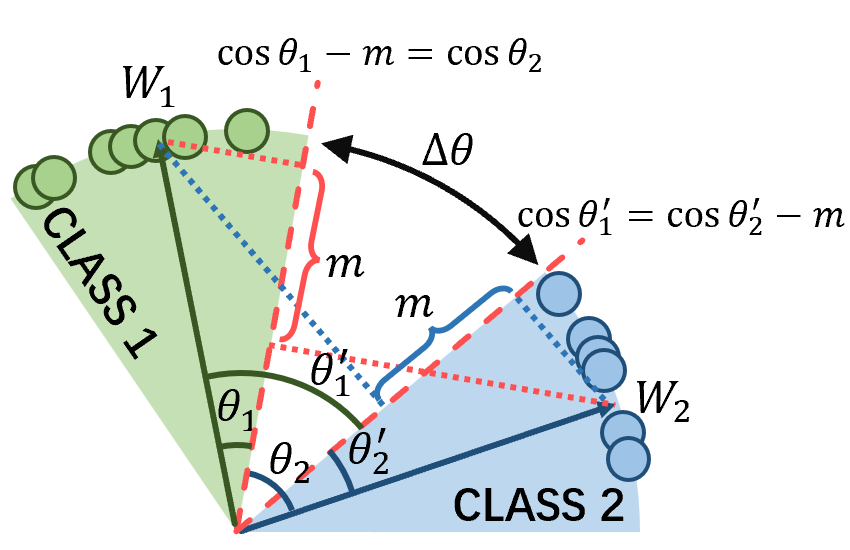}
            \label{sfig:dbCos}
        \end{minipage}
    }
    \subfigure[ArcFace]{
        \begin{minipage}[t]{0.3\linewidth}
             \centering
             \includegraphics[width=1\linewidth]{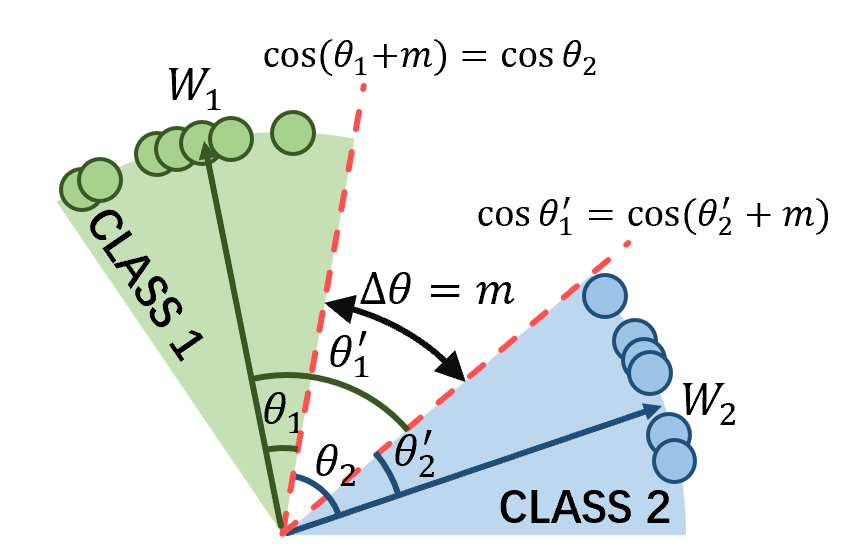}
            \label{sfig:dbArc}
        \end{minipage}
    }
    \caption{The decision boundaries and angular margins of different loss functions in binary classification problems are shown.}
    \label{fig:loss_desicion_boundary}
\end{figure}

For CosFace in a simple binary classification problem, suppose the face feature $x_{i}$ belongs to the first class. If $\cos\theta_{1}-m>\cos\theta_{2}$, then $x_{i}$ will be correctly classified as the first class.
If $\cos\theta_{1}-m<\cos\theta_{2}$, $x_{i}$ will be misclassified as the second class.
Thus, the decision boundary of CosFace is $\cos\theta_{1}-m=\cos\theta_{2}$. 
For ArcFace, when $\cos(\theta_{1}+m)>\cos\theta_{2}$, $x_{i}$ will be classified as the first class. When $\cos(\theta_{1}+m)<\cos\theta_{2}$, $x_{i}$ will be classified into second class.
Therefore, the decision boundary of ArcFace is $\cos(\theta_{1}+m)=\cos\theta_{2}$.
If the face feature $x_{i}$ belongs to the second class, the decision boundaries of CosFace and ArcFace are $\cos\theta_{1}=\cos\theta_{2}-m$ and $\cos\theta_{1}=\cos(\theta_{2}+m)$, respectively. 
Since the decision boundaries of CosFace and ArcFace do not overlap, there is an angular margin between classes, which ensures inter-class separation of the face features.
The angular margins of the above loss functions are shown in Fig.~\ref{fig:loss_desicion_boundary}.

For softmax loss, whose angular margin is zero because no matter which class $x_{i}$ belongs to, the decision boundary is always $W_{1}^{T}x_{i}=W_{2}^{T}x_{i}$.

For CosFace as shown in Fig.~\ref{sfig:dbCos}, we know that that $\theta_{1}+\theta_{2}=\theta'_{1}+\theta'_{2}\le\pi$, $\cos\theta_{1}-m=\cos\theta_{2}$ and $\cos\theta'_{1}=\cos\theta'_{2}-m$. 
Thus, we also know that $\theta_{1}=\theta'_{2}$ and $\theta_{2}=\theta'_{1}$.
Since $\cos\theta_{1}-m=\cos\theta_{2}$ and $\cos\theta'_{1}=\cos\theta'_{2}-m$, we have $\theta_{2}=\arccos(\cos\theta_{1}-m)$ and the margin between decision boundaries $\Delta\theta=\arccos(\cos\theta_{1}-m)-\theta_{1}$.

For ArcFace, the angular margin $\Delta\theta=\theta_{2}-\theta_{1}=\theta’_{1}-\theta’_{2}=m$.
Obviously, the hyperparameter $m$ of ArcFace determines the fixed angular margin.
Differences in logits functions of loss functions lead to different angular margins.

\subsection{X2-Softmax}
\label{subsec:x2s}
From the previous section, it can be seen that the angular margin between classes will affect the feature extraction. 
In this paper, we propose X2-Softmax loss which is margin adaptive.

The difference between our X2-Softmax, CosFace, and ArcFace is the logits function $f(\theta)$.
In our X2-Softmax loss, the logits function $f(\theta)$ is:
\begin{equation}
    \label{eq:x2sf}
    f_{R}(\theta)=a(\theta-h)^{2}+k
\end{equation}
where $a$, $h$, $k$ are hyperparameters of X2-Softmax loss.
$h$ and $k$ determine the vertice position of the logits function curve, and $a$ determines the direction of the opening of the curve and the degree of clustering. 

The cosine function is usually used as a logits function in traditional losses, e.g. CosFace and ArcFace.
In fact, the Taylor expansion of the cosine function is:
\begin{equation}
    \cos x=1-\frac{1}{2!}x^{2}+\frac{1}{4!}x^{4}+o(x^{4})
\end{equation}

Thus, converting the logits function to a quadratic function is a natural choice.
Discarding high-order terms of $x$ and retaining constant and quadratic terms can avoid the model overfitting while reducing computational effort.
By adding a primary term and changing the hyperparameters of the logits function, it is possible to make the margin more angle-adaptive and allow the logits curve to have more options for adjustments, which could enhance intra-class compactness and inter-class separation of face features.

Since the logits function is a quadratic function, we name the proposed loss function X2-Softmax.
X2-Softmax loss does not use a fixed angular margin, which makes it easier to determine the margin when faced with an imbalanced distribution of different samples.

\begin{figure}[ht]
    \centering
    \includegraphics[width=0.75\columnwidth]{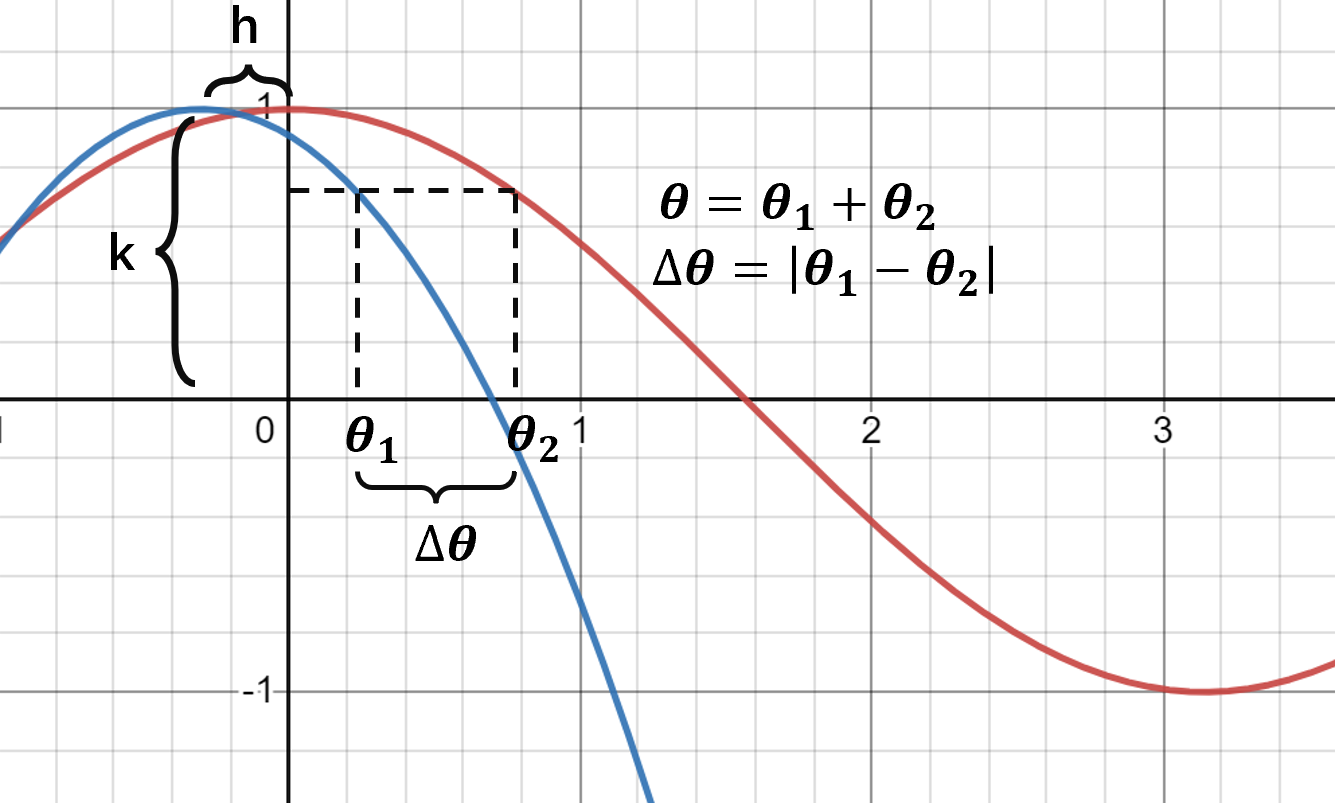}
    \caption{The curve of logits function $f(\theta)$ and the hyperparameters in X2-Softmax loss function.}
    \label{fig:x2sf_curve}
\end{figure}

Logits function's curve of X2-Softmax is shown in Fig.~\ref{fig:x2sf_curve}.

For the binary classification problem, when the face feature $x_{i}$ is classified into the first class, the decision boundary is $f_{R}(\theta_{1})=\cos(\theta_{2})$.
If the face feature $x_{i}$ is classified into the second class, the boundary is $f_{R}(\theta_{2})=\cos(\theta_{1})$.
This is shown in the Fig.~\ref{fig:x2sf_curve}, where the vertical coordinates of the two function curves always remain the same. 
Meanwhile, we assume that the angle between weights of two classes is $\theta=\theta_{1}+\theta_{2}$.
The angular margin between weights $\Delta\theta=\theta_{2}-\theta_{1}$.

From the decision boundary $f_{R}(\theta_{2})=\cos(\theta_{1})$ and Eq.~\ref{eq:x2sf}, we could get $\theta_{2}=\arccos(a(\theta_{1}-h)^{2}+k)$. 
Thus, the margin $\Delta\theta$ and the angle between the weights $\theta$ are given below:
\begin{equation}
    \label{eq:x2s_Dtheta}
    \Delta\theta=\arccos(a(\theta_{1}-h)^{2}+k)-\theta_{1}
\end{equation} 
\begin{equation}
    \label{eq:x2s_theta}
    \theta=\arccos(a(\theta_{1}-h)^{2}+k)+\theta_{1}
\end{equation}

According to Eq.~\ref{eq:x2s_Dtheta} and Eq.~\ref{eq:x2s_theta}, we plot the figure of the relation between the angle $\theta$ and the angular margin $\Delta\theta$ in~\ref{sfig:x2s_ang_mar}.

From Fig.~\ref{fig:x2sf_curve} and Fig.~\ref{sfig:x2s_ang_mar}, when the angle between weights $\theta$ increases, the angular margin $\Delta\theta$ monotonically increases at the same time.
The dynamic growing angular margin $\Delta\theta$ avoids the problem faced with a fixed angular margin and the larger margin can drive face features to cluster toward the weights.

The X2-Softmax loss function has set the margin in a way that is also more intuitive to us: two classes with a larger angle between weights would be set at a larger margin.

\subsection{Comparison with other loss}
In this section, we will explore the relationship between the angle of weights and the angular margin in different loss functions.

\begin{figure}[ht]
    \centering
    \subfigure[The curves of $\cos\theta$ and logits function $f(\theta)$ in ArcFace.]{
        \begin{minipage}[]{0.3\linewidth}
            \centering
            \includegraphics[width=1\linewidth]{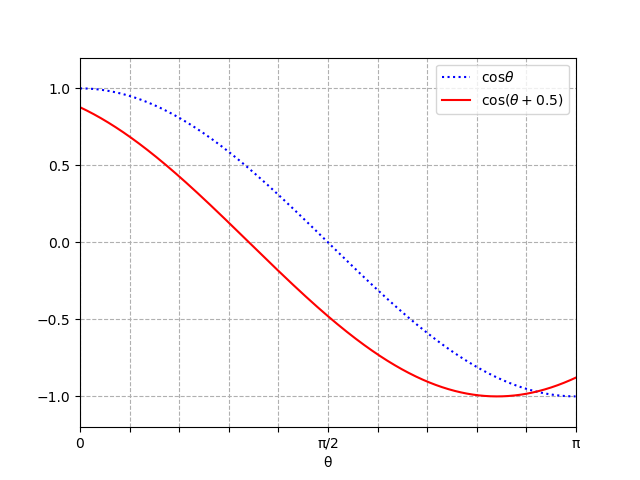}
        \end{minipage}
    }
    \subfigure[The curves of $\cos\theta$ and logits function $f(\theta)$ in CosFace.]{
        \begin{minipage}[]{0.3\linewidth}
            \centering
            \includegraphics[width=1\linewidth]{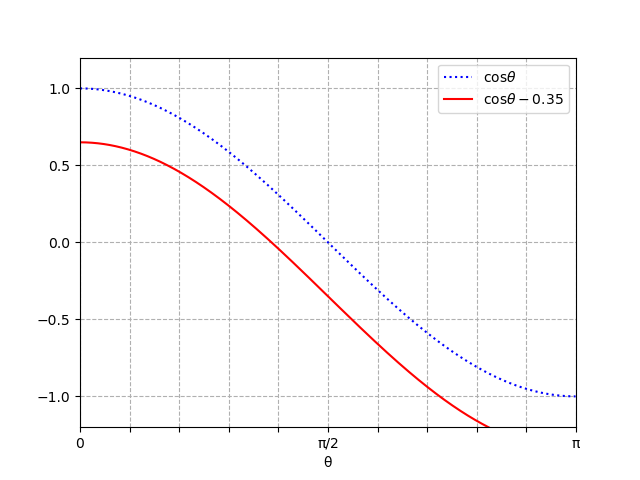}
        \end{minipage}
    }
    \subfigure[The curves of $\cos\theta$ and logits function $f(\theta)$ in X2-Softmax.]{
        \begin{minipage}[]{0.3\linewidth}
            \centering
            \includegraphics[width=1\linewidth]{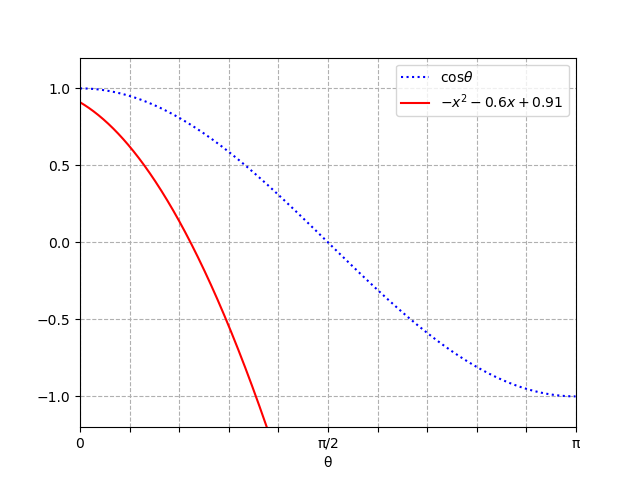}
        \end{minipage}
    }
    \subfigure[The relation between $\theta$ and $\Delta\theta$ in ArcFace.]{
        \begin{minipage}[]{0.3\linewidth}
            \centering
            \includegraphics[width=1\linewidth]{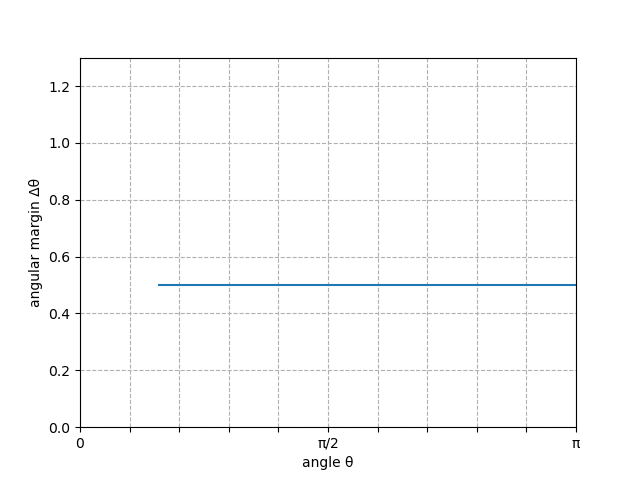}
            \label{sfig:arc_ang_mar}
        \end{minipage}
    }
    \subfigure[The relation between $\theta$ and $\Delta\theta$ in CosFace.]{
        \begin{minipage}[]{0.3\linewidth}
            \centering
            \includegraphics[width=1\linewidth]{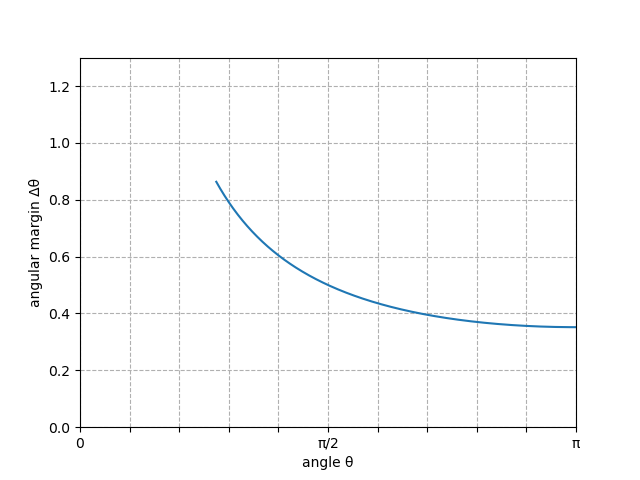}
            \label{sfig:cos_ang_mar}
        \end{minipage}
    }
    \subfigure[The relation between $\theta$ and $\Delta\theta$ in X2-Softmax.]{
        \begin{minipage}[]{0.3\linewidth}
            \centering
            \includegraphics[width=1\linewidth]{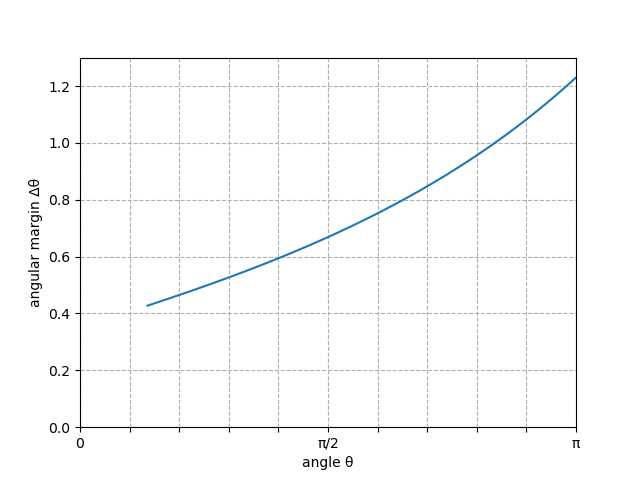}
            \label{sfig:x2s_ang_mar}
        \end{minipage}
    }
    \caption{The logits function and the relation between angle~$\theta$ and angular margin~$\Delta\theta$ in the different loss functions.
    For ArcFace, the angular margin is fixed and determined by hyperparameter $m$, no matter how large the angle is.
    For CosFace, the angular margin decreases while the angle increases, which does not make sense.
    In our X2-Softmax, the angular margin increases as the angle grows.}
    \label{fig:loss_curve}
\end{figure}

For ArcFace, since it applies a fixed additive angular margin directly in the angle space, the angular margin $\Delta\theta$ is determined by the hyperparameter $m$.
The margin would not change no matter what size of the angle between the weights of different classes.
However, ArcFace does not consider the actual similarity of two classes.
As shown in Fig.~\ref{sfig:arc_ang_mar}, it is clearly unreasonable that any two classes, regardless of how similar they are, always have the same size margin.

For CosFace, the angular margin between two classes is no longer a fixed value but varies with the angle between the weights of classes, because of its fixed margin applied in consine space.

However, noticed the formula of angular margin $\Delta\theta=\arccos(cos\theta_{1}+m)-\theta_{1}$ from Sec.~\ref{subsec:pre} and the angle between weights of classes $\theta=\arccos(cos\theta_{1}+m)+\theta_{1}$.
We plot the figure of the relation between $\theta$ and $\Delta\theta$ as Fig.~\ref{sfig:cos_ang_mar}.
AS $\theta$ increases, $\Delta\theta$ will decrease at the same time.
It means that the angular margin will instead be smaller for two classes less similar, which will reduce the inter-class separation of face features.
Whereas for two classes more similar, i.e. $\theta$ is smaller, the angular margin will be larger, which is more likely to make it difficult for the model to converge.

Both of these examples point out that there are some drawbacks in existing loss functions with a fixed angular margin.
Therefore, the X2-Softmax loss function considers that the angular margin should be adaptive and increase with the angle between weights growing.
As we mentioned at Sec~\ref{subsec:x2s}, Eq.~\ref{eq:x2s_Dtheta}, Eq.~\ref{eq:x2s_theta} and Fig.~\ref{sfig:x2s_ang_mar} show the relationship between angle $\theta$ and angular margin $\Delta\theta$.
The way $\Delta\theta$ increases with $\theta$ growing aligns with our intuition.
For two classes more similar, a suitable margin facilitates the model to complete convergence.
Whereas for two classes less similar, a larger margin will be assigned to enhance inter-class separation of face features.
Our X2-Softmax function is able to adaptively select angular margin.

The logits function $f(\theta)$ and relation between angle~$\theta$ and margin $\Delta\theta$ of the above loss function are shown in Fig.~\ref{fig:loss_curve}.

\subsection{Toy example}
To demonstrate the superiority of our X2-Softmax loss compared to other loss functions, we present a simple toy example.

All identities in MS1Mv3 with more than 120 images were selected and the face features of them are extracted.
Four identities with the highest variance and the other four identities with the lowest variance were selected in the toy example.
We trained the toy example on ResNet-34 for these eight identities and mapped them into a 2D face feature space.  

\begin{figure}[ht]
    \centering
    \subfigure[ArcFace(s=64, m=0.5)]{
        \begin{minipage}[]{0.3\linewidth}
            \centering
            \includegraphics[width=1\linewidth]{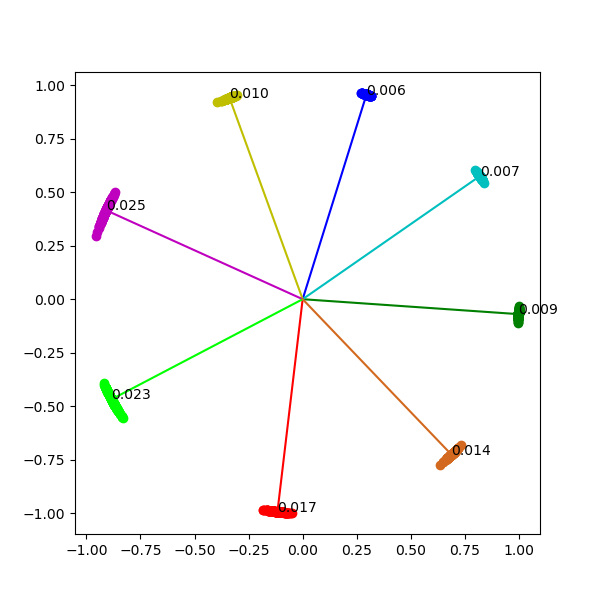}
        \end{minipage}
    }
    \subfigure[ElasticFace-Arc(s=64, m=0.5, $\sigma$=0.05)]{
        \begin{minipage}[]{0.3\linewidth}
            \centering
            \includegraphics[width=1\linewidth]{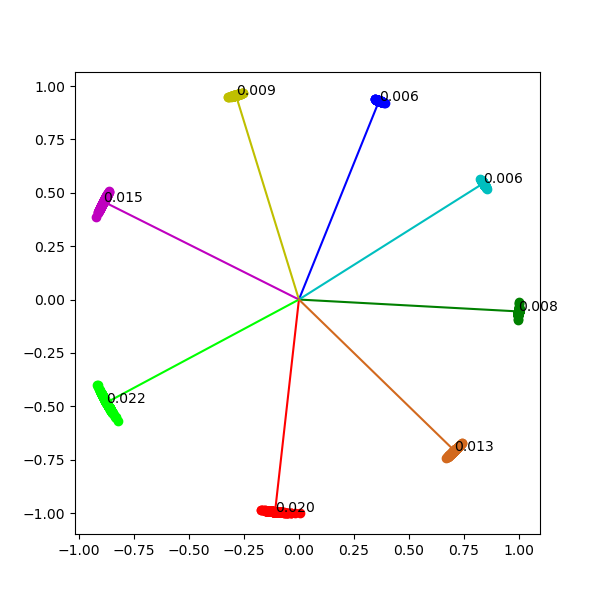}
        \end{minipage}
    }
    \subfigure[X2-Softmax(s=64, a=-1, h=0.3, k=1)]{
        \begin{minipage}[]{0.3\linewidth}
            \centering
            \includegraphics[width=1\linewidth]{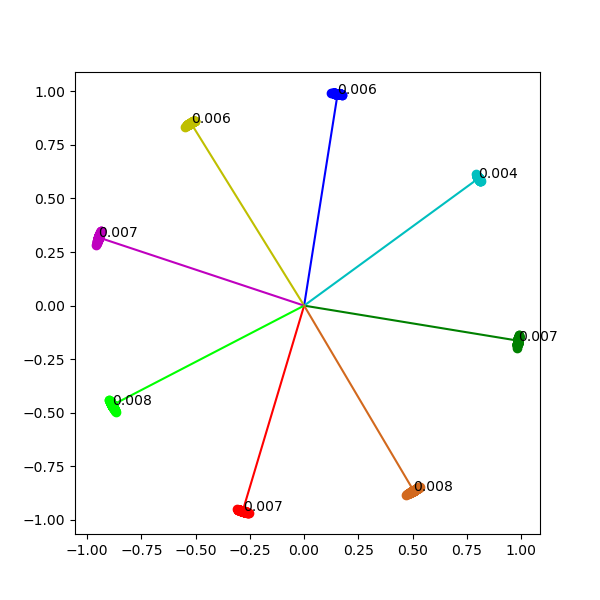}
        \end{minipage}
    }
    \caption{Toy examples of networks trained with different loss functions are shown.
    The number on each identity indicates the standard deviation of angles between the mean face feature of identity to every face feature, which indicates the intra-class compactness of an identity.}
    \label{fig:toyexample}
\end{figure}

As shown in Fig.~\ref{fig:toyexample}face images from eight identities are mapped into a 2D face feature space after training them with different loss functions.
The standard deviations of angles between features to the mean features are labeled in the figure.
From the results, it can be seen that the standard deviation of X2-Softmax is small, which means the face features extracted by the model trained with X2-Softmax are more compact in each class than the others. 

\section{Experiments}
\subsection{Implementation Details}
\textbf{Experimental settings.} 
The experiments in this paper were implemented by Pytorch~\cite{paszke2019pytorch}. 
We experimented with the Windows 10 system and used NVIDIA GeForce RTX 3090 for training. 

The neural network models used to demonstrate all experimental training and testing in this paper are Resnet50~\cite{he2016deep}, which is widely used in face recognition solutions~\cite{deng2019arcface, boutros2022elasticface}. 
The models are trained by a Stochastic Gradient Descent (SGD) optimizer with an initial learning rate of 0.02.
Momentum and weight decay are set to 0.9 and 5e-4, respectively, and the batch size is set to 128. 
The total number of training iterations is 1052k. 
Learning rate is multiplied by 0.1 at 323k, 566k, 809k, and 1011k training iterations. 
Data augmentation is performed by random horizontal flipping with a probability of 0.5 during training. 
The size of images used for training and evaluation is ~$112\times112\times3$, and the dimension of the face features is 512.
All pixel values of images used for training and evaluation are normalized to between -1 and 1. 
As the common setting, we set the hyperparameter $s$ to 64.

\textbf{Training datasets.} 
The training set used for experiments in this paper is MS1Mv3~\cite{deng2021masked}, which has been widely used in recent years in face recognition solutions~\cite{deng2021masked, zhang2022towards}. 
The MS1Mv3 is based on the MS-Celeb-1M dataset~\cite{guo2016ms}, which contains about 10M images and 100k identities. 
Since the original version has a lot of noisy images, the refined version is preprocessed by RetinaFace~\cite{deng2019retinaface}. 
After removing the noisy image, MS1Mv3 includes 93k identities and 5.1M face images.

\textbf{Evaluation benchmarks.} 
In order to demonstrate the effectiveness of X2-Softmax loss on face recognition and compare it with other loss functions, we evaluate the trained models on eight different evaluation benchmarks. 
These eight evaluation benchmarks are 
1) Labeled Faces in the Wild (LFW)~\cite{huang2008labeled}, 
2) AgeDB-30~\cite{moschoglou2017agedb}, 
3) Cross-age LFW (CALFW)~\cite{zheng2017cross}, 
4) Cross-Pose LFW (CPLFW)~\cite{zheng2018cross}, 
5) Celebrities in Frontal-Profile in the Wild (CFP-FP)~\cite{sengupta2016frontal}, 
6) Visual Geometry Group Face 2 in Frontal-Profile (VGGFace2-FP)~\cite{cao2017vggface2}, 
7) IARPA Janus Benchmark-B (IJB-B)~\cite{whitelam2017iarpa}, and 
8) IARPA Janus Benchmark-C (IJB-C)~\cite{maze2018iarpa}. 
For the first six evaluation benchmarks, we report the accuracies of models trained with different loss functions to evaluate the effectiveness. 
For the latter two evaluation benchmarks IJB-B and IJB-C, we calculated the metric called True Accept Rate at False Accept Rate (TAR@FAR), where FAR is set to 1e-4 and 1e-5 respectively. 
The number of images and identities in different face datasets used for training and evaluation are shown in Table~\ref{tb:dataset}.

\begin{table}[h]
\centering
\begin{tabular}{l l l}
\toprule
Datasets & Identity & Image \\ \midrule
MS1Mv3~\cite{deng2021masked}   & 93k      & 5.1M   \\ \midrule
LFW~\cite{huang2008labeled}     & 5749     & 13233  \\
AgeDB-30~\cite{moschoglou2017agedb} & 568      & 16488  \\
CALFW~\cite{zheng2017cross}   & 5749     & 11652  \\
CPLFW~\cite{zheng2018cross}   & 5749     & 12174  \\
CFP-FP~\cite{sengupta2016frontal}  & 500      & 7000   \\
VGG2-FP~\cite{cao2017vggface2} & 9131     & 3.31M  \\ \midrule
IJB-B~\cite{whitelam2017iarpa}   & 1845     & 76.8k  \\
IJB-C~\cite{maze2018iarpa}   & 3531     & 148.8k \\ \bottomrule
\end{tabular}
\caption{The numbers of identities and images in the training set and evaluation benchmarks used in the experiment.
}
\label{tb:dataset}
\end{table}

\subsection{Hyperparameter study} 
In this section, we will perform a hyperparameter study to investigate the effect of different hyperparameters in X2-Softmax loss. 

The hyperparameters $a$, $h$, and $k$ together determine the logits function curve in X2-Softmax loss and the difference between the logits function curve and $\cos\theta$.
Hyperparameter $a$ determines the direction of the opening and the degree of convergence of the logits function curve. 
The logits function should decrease as the angle between the face feature~$x_{i}$ and the weight~$W_{y_{i}}$ increases, so the hyperparameter $a$ should be set to a negative number.
As the absolute value of $a$ increases, the logits function curve becomes more clustered and steeper.
The angular margin increases as the absolute value of $a$ increases.
For two classes with higher similarity and a smaller weights angle, the increase in angular margin by $a$ increasing is larger than the increase in margin for the two classes with lower similarity and larger weights angle.

Hyperparameter $h$ represents the horizontal coordinate of the vertex of the logits function curve. As the hyperparameter $h$ decreases, the logits function curve shifts to the left and the difference between the logits function curve and $\cos\theta$ function curve increases, which means the angular margin increases meanwhile.
Unlike the effect of margin with changing in $a$, the effect of angular margin with changing in $h$ is independent of the similarity and angle between two classes.

\begin{figure}[ht]
    \centering
    \subfigure[Hyperparameter $a$.]{
        \begin{minipage}[]{0.3\linewidth}
            \centering
            \includegraphics[width=1\linewidth]{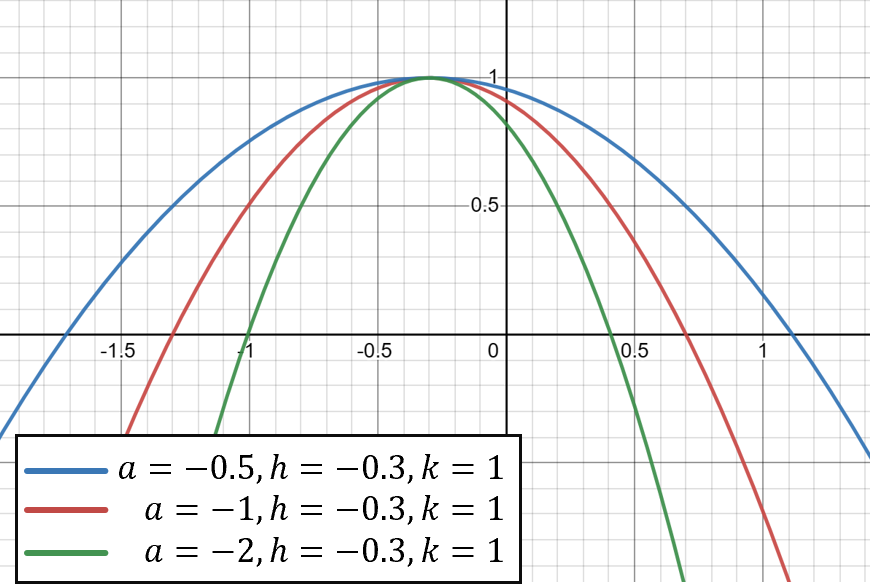}
        \end{minipage}
    }
    \subfigure[Hyperparameter $h$.]{
        \begin{minipage}[]{0.3\linewidth}
            \centering
            \includegraphics[width=1\linewidth]{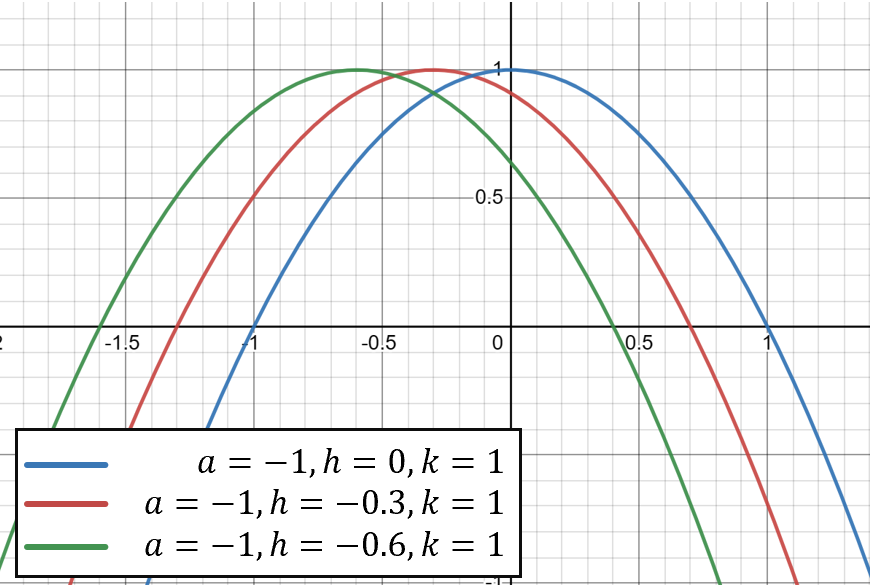}
        \end{minipage}
    }
    \subfigure[Hyperparameter $k$.]{
        \begin{minipage}[]{0.3\linewidth}
            \centering
            \includegraphics[width=1\linewidth]{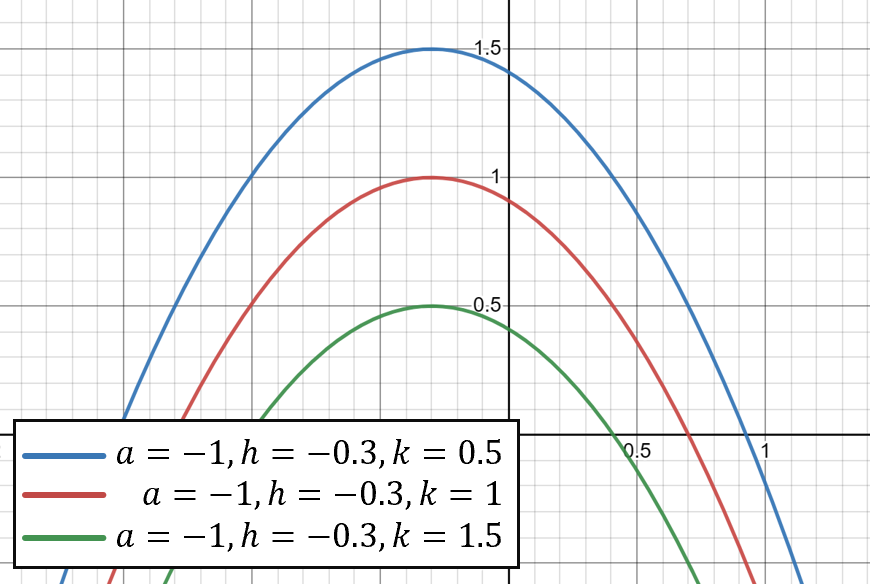}
        \end{minipage}
    }
    \caption{Hyperparameters $a$, $h$, and $k$ determine the curve of the logits function in X2-Softmax loss.}
    \label{fig:hyperparameter}
\end{figure}

Hyperparameter $k$ represents the vertical coordinate of the vertex of the logits function curve.
The angular margin increases with hyperparameter $k$ decreasing.
Since the three hyperparameters $a$, $h$, and $k$ affect the logits function curve and the angular margins between classes, we selected different value cases of these three hyperparameters for the parameterization experiments.
As shown in Fig.~\ref{fig:hyperparameter}, the hyperparameters determine the curve of the logits function in X2-Softmax.

The experiment results are shown in Table~\ref{tab:para}.
The trained model achieved the best synthesis result when the hyperparameter was chosen as $a=-1.0$, $h=-0.3$, and $k=1.0$.

\begin{table*}[h]
\centering
\begin{tabular}{@{}lll|ccc@{}}
\toprule
\multirow{2}{*}{} &  &  & LFW & AgeDB-30 & CPLFW \\
&  &  & Acc(\%) & Acc(\%) & Acc(\%) \\ \midrule
a=-0.7 & h=-0.3 & k=1.0 & 99.78 & 96.13 & 92.12 \\
a=-0.9 & h=-0.3 & k=1.0 & 99.78 & 95.95 & 92.12 \\
a=-1.0 & h=-0.3 & k=1.0 & 99.82 & 95.92 & 91.67 \\
a=-1.1 & h=-0.3 & k=1.0 & 99.82 & 96.00 & 92.07 \\
a=-1.3 & h=-0.3 & k=1.0 & 99.82 & 95.98 & 92.22 \\ \midrule
a=-1.0 & h=-0.5 & k=1.0 & 99.80 & 95.90 & 91.93 \\
a=-1.0 & h=-0.4 & k=1.0 & 99.78 & 95.97 & 92.27 \\
a=-1.0 & h=-0.2 & k=1.0 & 99.78 & 96.05 & 92.27 \\
a=-1.0 & h=-0.1 & k=1.0 & 99.82 & 96.12 & 92.27 \\
a=-1.0 & h=0.0 & k=1.0 & 99.78 & 95.90 & 92.17 \\ \midrule
a=-1.0 & h=-0.3 & k=0.9 & 99.77 & 95.98 & 92.33 \\
a=-1.0 & h=-0.3 & k=0.8 & 99.82 & 95.87 & 91.93 \\
a=-1.0 & h=-0.3 & k=0.7 & 99.80 & 96.07 & 92.25 \\ \bottomrule
\end{tabular}
\caption{Hyperparameter study experiment results.}
\label{tab:para}
\end{table*}

\begin{table*}[ht]
\centering
\begin{tabular}{@{}l|llllll@{}}
\toprule
\multirow{2}{*}{Method} & LFW & CALFW & CPLFW & AgeDB-30 & CFP-FP & VGG2-FP \\
& Acc(\%) & Acc(\%) & Acc(\%) & Acc(\%) & Acc(\%) & Acc(\%) \\ \midrule
Softmax & 99.67 & 95.22 & 90.48 & 96.60 & 96.99 & \textbf{95.56} \\
NormFace~\cite{wang2017normface} & 99.40 & 93.73 & 88.68 & 94.25 & 94.71 & 9318 \\
CosFace~\cite{wang2018cosface} & 99.75 & 96.02 & 92.35 & 97.90 & \textbf{98.01} & 95.24 \\
ArcFace~\cite{deng2019arcface} & 99.77 & 96.03 & 92.05 & \textbf{98.18} & 97.59 & 95.30 \\
Dyn-ArcFace~\cite{jiao2021dyn} & 99.82 & 95.97 & 92.35 & 97.97 & 97.69 & 94.96 \\
MagFace~\cite{meng2021magface} & 99.82 & 96.07 & 92.23 & 97.93 & 97.90 & 95.14 \\
ElasticFace-Arc~\cite{boutros2022elasticface} & 99.80 & \textbf{96.12} & \textbf{92.48} & 98.02 & 97.61 & 95.06 \\
ElasticFace-Cos~\cite{boutros2022elasticface} & 99.80 & 96.07 & 92.42 & 97.98 & 97.69 & 94.88 \\
X2-Softmax & \textbf{99.82} & 95.92 & 91.67 & 97.83 & 97.20 & 94.52 \\ \bottomrule
\end{tabular}
\caption{The accuracies of different loss functions on six evaluation benchmarks, LFW, CALFW, CPLFW, AgeDB-30, CFP-FP, and VGG2-FP are shown.
The highest accuracy on each evaluation benchmark is shown in bold.}
\label{tb:resultsimple}
\end{table*}

\subsection{Results with Different Loss Functions} 
In this section, we will report the results of model training with X2-Softmax loss on eight evaluation benchmarks, and compare them with the results of other loss functions.

\textbf{Results on LFW, CALFW, CPLFW, AgeDB-30, CFP-FP, VGG2-FP.} 
Table~\ref{tb:resultsimple} shows the accuracy of the models training with different loss functions on LFW~\cite{huang2008labeled}, CALFW~\cite{zheng2017cross}, CPLFW~\cite{zheng2018cross}, AgeDB-30~\cite{moschoglou2017agedb}, CFP-FP~\cite{sengupta2016frontal}, VGG2-FP~\cite{cao2017vggface2}.

LFW benchmark is most widely used in face recognition programs, and it is also one of the oldest benchmarks. 
It is nearing saturation in terms of accuracy.
The differences in the results of the different evaluation benchmarks mainly come from different face variations, such as different ages or poses, e.g., CALFW and CPLFW. 
On LFW benchmark, we achieved the best results with an accuracy of 99.82\%.
On CALFW and CPLFW benchmarks, our X2-Softmax still achieves the accuracies of 95.92\% and 91.67\%, respectively, which are close to optimal accuracies. 
On AgeDB-30, the accuracy of X2-Softmax is 97.83\%.
In the CFP-FP benchmark, the model has to match frontal and side faces, which undoubtedly increases the difficulty of face recognition. 
In this benchmark, CosFace achieves optimal results with an accuracy of 98.01\%, and X2-Softmax loss achieves an accuracy of 97.20\%.
On VGG2-FP benchmark, the accuracy of X2-Softmax is 94.52\%.
The results are close to saturation on each of these evaluation benchmarks.
The accuracies of different loss functions are also close to each other, thus it is not sufficient to evaluate by these few evaluation benchmarks.
Therefore, we also used the IJB evaluation benchmarks to compare X2-Softmax with loss functions. 

\textbf{Results on IJB-B, IJB-C.} 
IJB-B and IJB-C are more important evaluation benchmarks for face recognition in recent years.
The results for IJB-B and IJB-C benchmarks are shown in Table~\ref{tb:resultijb}.
As shown in the table, our X2-Softmax loss leads in three of the four evaluation benchmarks.

On the IJB-B benchmark, our X2-Softmax reaches TAR of 0.9495 at FAR equal to 1e-4, and at FAR equal to 1e-5, the TAR of X2-Softmax reaches 0.9133, both ahead of other state-of-the-art loss functions. 
On the IJB-C benchmark, when FAR is equal to 1e-4, X2-Softmax reaches TAR of 0.9624, slightly behind other loss functions
When FAR is equal to 1e-5, the TAR of X2-Softmax loss reaches 0.9459, still ahead of other loss functions.

Both on the IJB-B and IJB-C benchmarks, our X2-Softmax performs well and most of the time is able to outperform other loss functions.  
It is clear that our X2-Softmax loss function improves performance on both IJB-B and IJB-C.

\begin{table*}[h]
\centering
\label{tb:resultijb}
\begin{tabular}{l|ll|ll}
\toprule
\multirow{2}{*}{Method} & \multicolumn{2}{c|}{IJB-B} & \multicolumn{2}{c}{IJB-C} \\ %\cline{2-5} 
 & \multicolumn{1}{l}{\begin{tabular}[c]{@{}l@{}}TAR \\@ FAR = 1e-4\end{tabular}} & \begin{tabular}[c]{@{}l@{}}TAR \\@ FAR = 1e-5\end{tabular} & \multicolumn{1}{l}{\begin{tabular}[c]{@{}l@{}}TAR \\@ FAR = 1e-4\end{tabular}} & \begin{tabular}[c]{@{}l@{}}TAR \\@ FAR = 1e-5\end{tabular} \\ \midrule
Softmax & \multicolumn{1}{l}{0.8762}  & 0.7501 & \multicolumn{1}{l}{0.8982} & 0.8012 \\
NormFace & \multicolumn{1}{l}{0.8377} & 0.7233 & \multicolumn{1}{l}{0.8757} & 0.8016 \\
CosFace & \multicolumn{1}{l}{0.9456} & 0.8959 & \multicolumn{1}{l}{0.9606} & 0.9418 \\
ArcFace & \multicolumn{1}{l}{0.9485} & 0.9092 & \multicolumn{1}{l}{0.9629} & 0.9449 \\
Dyn-ArcFace & \multicolumn{1}{l}{0.9490} & 0.9124 & \multicolumn{1}{l}{0.9627} & 0.9447 \\
MagFace & \multicolumn{1}{l}{0.9380} & 0.8745 & \multicolumn{1}{l}{0.9555} & 0.9301 \\
ElasticFace-Arc & \multicolumn{1}{l}{0.9487} & 0.9118 & \multicolumn{1}{l}{\textbf{0.9632}} & 0.9457 \\
ElasticFace-Cos & \multicolumn{1}{l}{0.9473} & 0.9021 & \multicolumn{1}{l}{0.9622} & 0.9446 \\
X2-Softmax & \multicolumn{1}{l}{\textbf{0.9495}} & \textbf{0.9133} & \multicolumn{1}{l}{0.9624} & \textbf{0.9459} \\ \bottomrule
\end{tabular}
\caption{
The table shows the results for different loss functions on the IJB-B and IJB-C benchmarks. The highest TAR at different FARs on each benchmark is shown in bold.
}
\end{table*}

\begin{figure}[ht]
    \centering
    \subfigure[ArcFace]{
        \begin{minipage}[]{0.3\linewidth}
            \centering
            \includegraphics[width=1\linewidth]{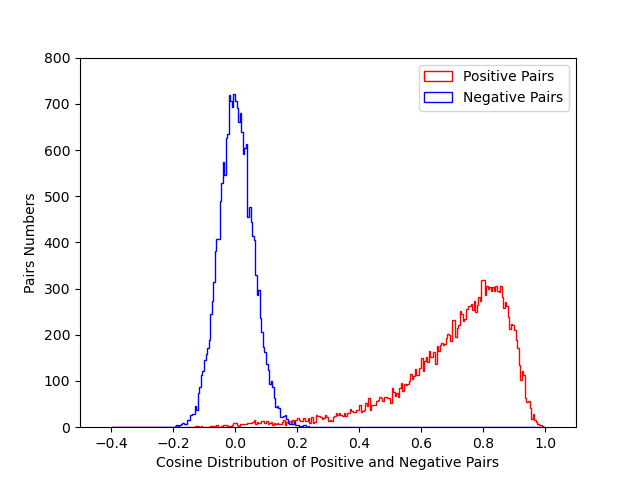}
        \end{minipage}
    }
    \subfigure[Elastic-Arc]{
        \begin{minipage}[]{0.3\linewidth}
            \centering
            \includegraphics[width=1\linewidth]{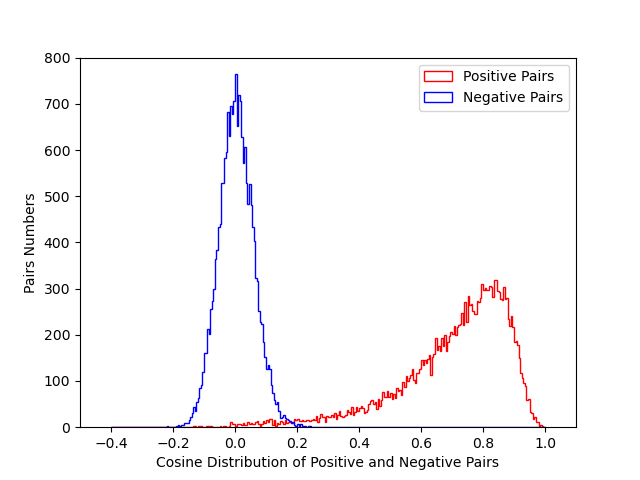}
        \end{minipage}
    }
    \subfigure[X2-Softmax]{
        \begin{minipage}[]{0.3\linewidth}
            \centering
            \includegraphics[width=1\linewidth]{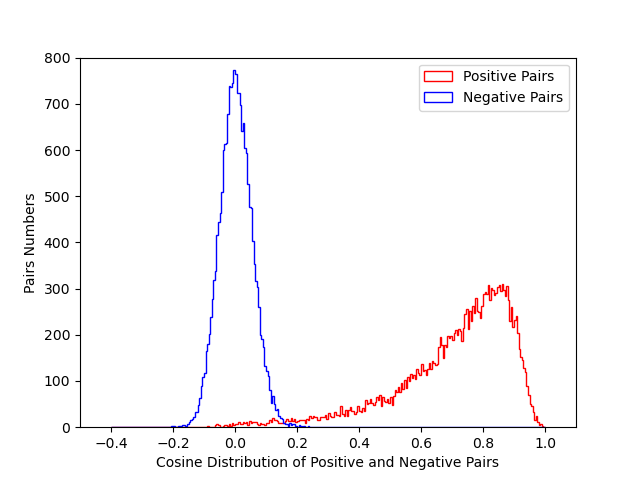}
        \end{minipage}
    }
    \caption{Cosine distribution of positive and negative sample pairs features in ijbc dataset, extracted by models trained with different loss functions.}
    \label{fig:cos_sim_ijbc}
\end{figure}

\textbf{Cosine similarity distribution of IJB-C positive and negative sample pairs extracted by different loss functions.} 
Fig.~\ref{fig:cos_sim_ijbc} represents the cosine values of positive and random negative sample pairs for the IJB-C evaluation benchmark with the network architecture of ResNet-50 trained by different loss functions. 
The red curves of the figure represent the distribution of cosine values for all positive sample pairs (about 10k) in IJB-C, and the blue curves are the distribution of cosine values for randomly selected negative sample pairs equal to the number of positive sample pairs. 
The red and blue overlapping parts of the figure represent the confusion region of positive and negative sample pairs. 
From the results, the confusion region of our X2-Softmax is smaller, indicating that this loss function separates face features more thoroughly.

\section{Conclusion}
This paper proposes a simple yet effective and efficient loss function called X2-Softmax loss with adaptive margins for face recognition. 
Specifically, due to the margin adaptivity of this loss, it is able to apply an adaptive margin according to the angle between two classes during training, which increases with the weight angle growing.
It also avoids searching for a suitable hyperparameter $m$ as the margin.
We have tested it on several datasets. The experimental results demonstrate the effectiveness and superiority of our method, the performance of which is promising on difficult benchmarks.

\section*{Acknowledgement}
This work was supported in part by the Guangdong Basic and Applied Basic Research Foundation under Grant 2022A515110020, the Guangzhou Basic and Applied Basic Research Foundation under Grant 202201010476, and the National Natural Science Foundation of China under Grant 62271232.




\bibliographystyle{elsarticle-num} 
\bibliography{main}


%\begin{thebibliography}{00}


%\bibitem{}

%\end{thebibliography}
\end{document}